%% file: main.tex
\documentclass[runningheads]{llncs}
\usepackage{pifont}

\input{macros}

\usepackage{eccv}

\usepackage{eccvabbrv}
\usepackage{subcaption}
\usepackage{graphicx}
\usepackage{booktabs}
\usepackage[accsupp]{axessibility}  

\usepackage{hyperref}

\def\Ours{Human-LRM}

\usepackage{orcidlink}

\begin{document}

\title{Template-Free Single-View 3D Human Digitalization with Diffusion-Guided LRM}

\titlerunning{Human-LRM}

\author{Zhenzhen Weng$^{1}$, Jingyuan Liu\textsuperscript{2}, Hao Tan\textsuperscript{2}, Zhan Xu\textsuperscript{2}, Yang Zhou\textsuperscript{2}\\ 
Serena Yeung-Levy\textsuperscript{1}, Jimei Yang\textsuperscript{2}
}

\authorrunning{Weng et al.}

\institute{
\textsuperscript{1}Stanford University, \textsuperscript{2}Adobe Research \\
\email{{\tt\small \textsuperscript{1}\{zzweng,syyeung\}@stanford.edu, \textsuperscript{2}\{jingyliu,hatan,zhaxu,yazhou,jimyang\}@adobe.com}
}
}


\maketitle

\begin{center}
    \centering
    \captionsetup{type=figure}
    \includegraphics[width=\textwidth]{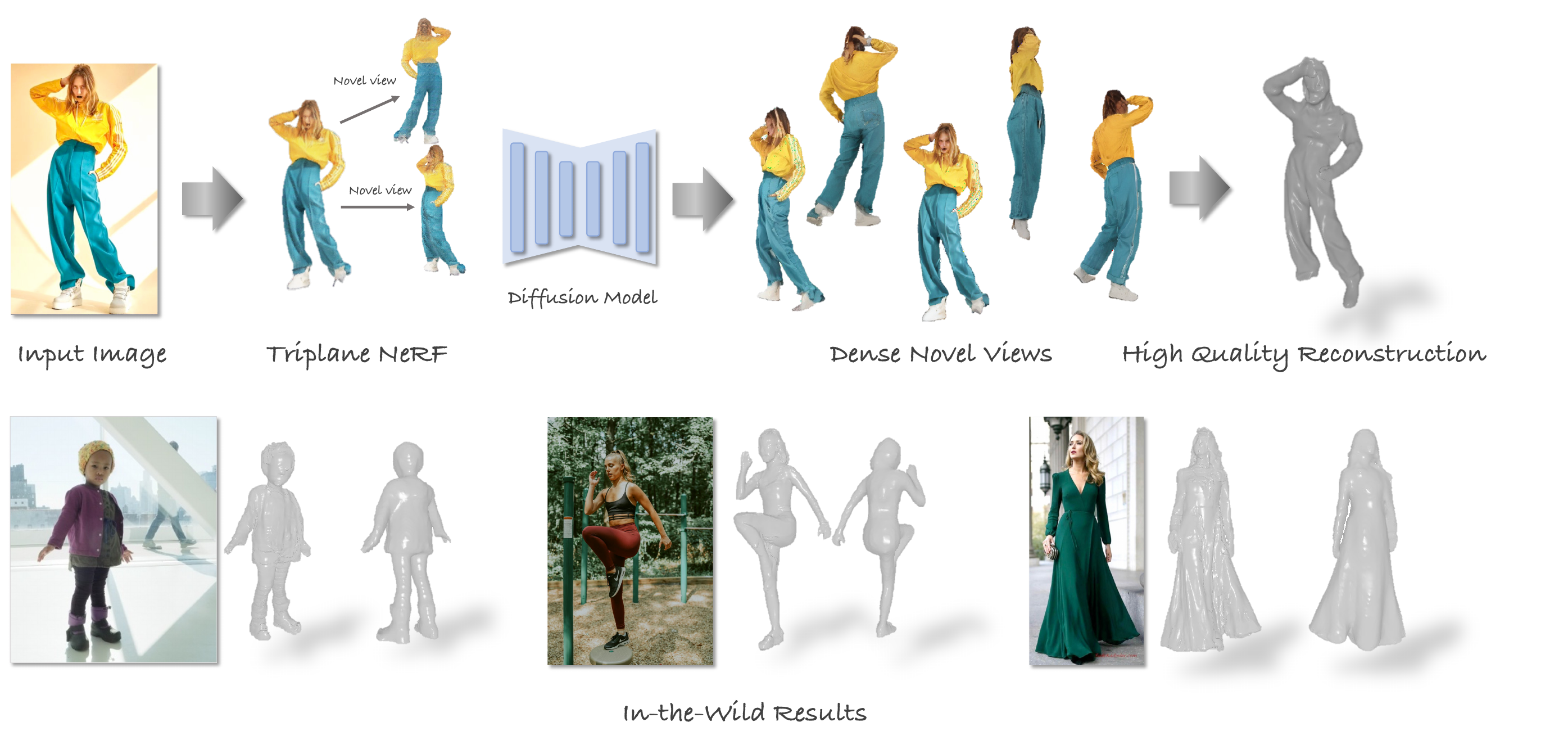}
    \captionof{figure}{
    We present \Ours, a template-free large reconstruction model for feed-forward 3D human digitalization from a single image. Trained on a vast dataset comprising multi-view capture and 3D scans, our model generalizes across a broader range of scenarios. Guided by dense novel views generated by a conditional diffusion model, our model can generate high-fidelity full body humans from a single image. Our project webpage is at~\url{https://zzweng.github.io/humanlrm}.
    }
    \label{fig:pull}
\end{center}

\input{sec/0_abstract}    
\input{sec/1_intro}
\input{sec/2_related_work}

\input{sec/3_method}
\input{sec/4_experiments}
\input{sec/5_conclusion}
\input{sec/6_acknowledgement}
\input{sec/X_suppl}

%
%
\bibliographystyle{splncs04}
\bibliography{main}
\end{document}

%% file: macros.tex
\newcommand{\bc}{\mathbf{c}}
\newcommand{\bd}{\mathbf{d}}

\newcommand{\bh}{\mathbf{h}}

\newcommand{\bn}{\mathbf{n}}
\newcommand{\bo}{\mathbf{o}}

\newcommand{\br}{\mathbf{r}}

\newcommand{\bv}{\mathbf{v}}
\newcommand{\bw}{\mathbf{w}}
\newcommand{\bx}{\mathbf{x}}

\newcommand{\bT}{\mathbf{T}}

\def\D{{\cal D}}
\def\E{{\cal E}}

%% file: sec/0_abstract.tex
\begin{abstract}



Reconstructing 3D humans from a single image has been extensively investigated. However, existing approaches often fall short on capturing fine geometry and appearance details, hallucinating occluded parts with plausible details, and achieving generalization across unseen and in-the-wild datasets.
We present \Ours{}, a diffusion-guided feed-forward model that predicts the implicit field of a human from a single image. Leveraging the power of the state-of-the-art reconstruction model (i.e., LRM) and generative model (i.e Stable Diffusion), our method is able to capture human without any template prior, e.g., SMPL, and effectively enhance occluded parts with rich and realistic details.
Our approach first uses a single-view LRM model with an enhanced geometry decoder to get the triplane NeRF representation. The novel view renderings from the triplane NeRF provide strong geometry and color prior, from which we generate photo-realistic details for the occluded parts using a diffusion model. The generated multiple views then enable reconstruction with high-quality geometry and appearance, leading to superior overall performance comparing to all existing human reconstruction methods.

\end{abstract}

%% file: sec/1_intro.tex
\section{Introduction}
\label{sec:intro}
\begin{figure}[h]
    \centering
    \includegraphics[width=\columnwidth]{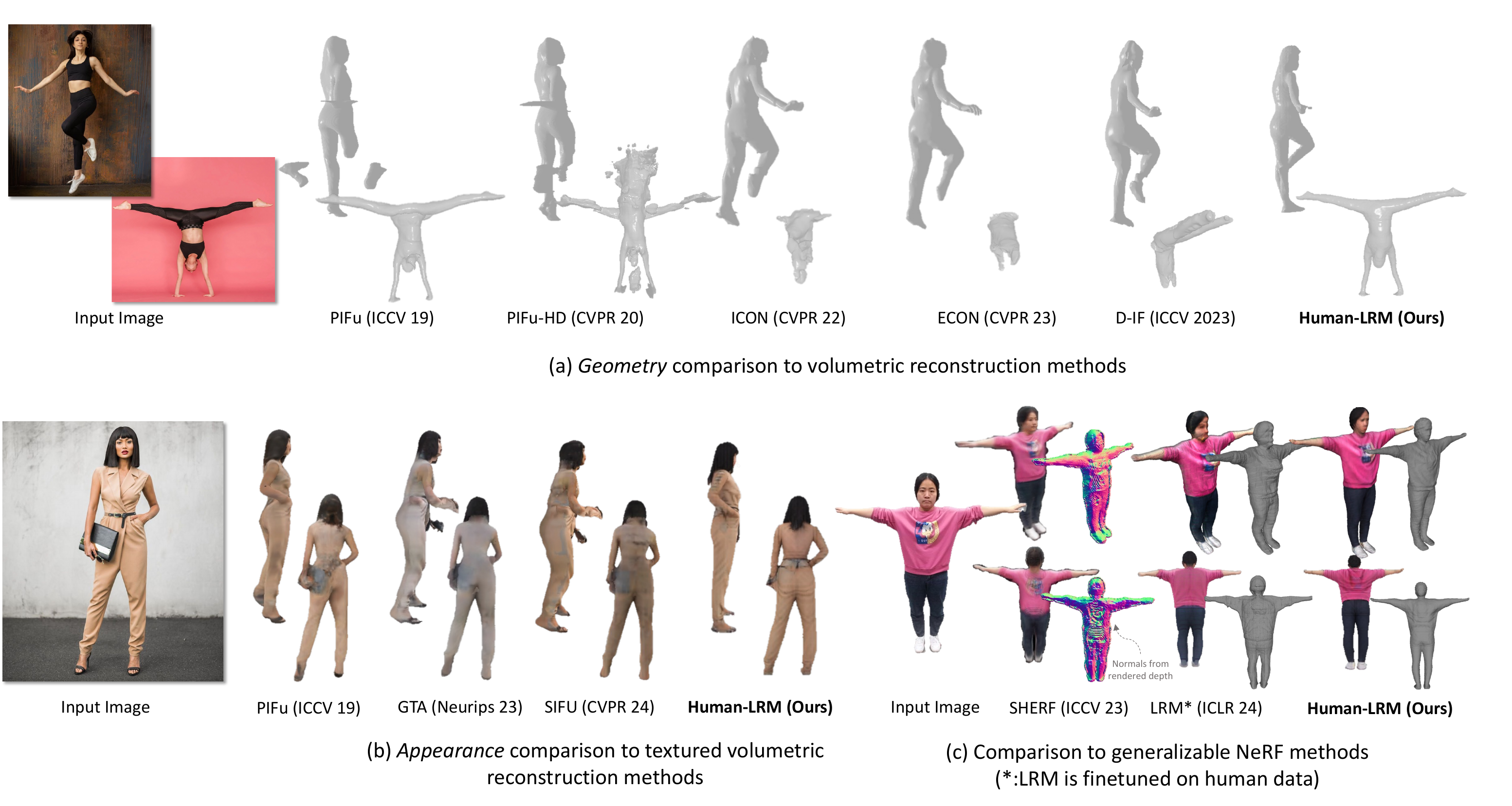}
    \caption{
    Comparison of \Ours{} with SoTA single-view human reconstruction methods on in-the-wild images. Compared to volumetric reconstruction methods, our method achieves superior generalizability to challenging poses (a) and higher fidelity appearance prediction (b). Compared to generalizable human NeRF methods (c), our result achieves much better geometry quality.
    }
    \label{fig:intro_fig}
\end{figure}

Reconstructing 3D human models from a single image is an important research topic in computer vision with an array of practical applications. These applications encompass areas such as AR/VR, asset creation, relighting, and many more. A plethora of techniques have been developed to address this challenging task, each with its own set of advantages and limitations. Parametric reconstruction methods, a.k.a. human mesh recovery (HMR) \cite{kanazawa2018end,feng2021collaborative,zhang2021pymaf} regress pose and shape parameters of SMPL (Skinned Multi-Person Linear) human body mesh model \cite{loper2023smpl}, which does not include clothing details. This limits their utility in applications requiring realistic and detailed human representations. Conversely, implicit volume reconstruction methods \cite{saito2019pifu,saito2020pifuhd} 
capture fine-grained clothing details with their pixel-aligned features but do not generalize across various poses.
Recent hybrid approaches \cite{xiu2022icon,xiu2023econ,zheng2021pamir,yang2023d,zhang2024global,zhang2023sifu} combine the advantages of parametric and implicit reconstruction methods by using the predicted SMPL body mesh as conditioning to guide the full clothed reconstruction. 
However, these SMPL-conditioned methods face inevitable limitations: SMPL prediction errors propagate to the subsequent full reconstruction stage, resulting in misalignment between the reconstructed mesh and the input image. The misalignment is exacerbated when the poses are complex (Fig.~\ref{fig:intro_fig} (a)). These errors are often irreparable and cannot be fully fixed by post-hoc optimization \cite{xiu2022icon,zheng2021pamir,xiu2023econ}. Moreover, these works typically do not learn the appearances. For the works that perform joint prediction of geometry and appearances, the appearance predictions are blurry especially on the occluded part (Fig.~\ref{fig:intro_fig} (b)).

Meanwhile, there have been various works that use NeRF \cite{mildenhall2021nerf} as a representation to learn geometry as well as texture of humans, but these works typically overfit to single scenes~\cite{huang2023tech,albahar2023humansgd}, which is not generalizable to new observations. Recently, feed-forward NeRF prediction models such as Large Reconstruction Model (LRM) \cite{hong2023lrm} has been proposed, which enables generalizable 3D reconstructions from arbitrary single image inputs. However, directly applying LRM to humans yields sub-optimal results even with fine-tuning (Fig.~\ref{fig:intro_fig} (c)). Primarily, the reconstructed surfaces tend to be coarse, not preserving enough details. Second, the occluded part has collapsed appearances and appear blurry.


In this work, we present \Ours{}, a feed-forward model that predicts the geometry and appearance of the human from a single image. We draw from the insight that diffusion models yield high-quality, novel view hallucinations for occluded parts, while 3D reconstruction provide strong geometry and color prior to ensure the diffusion model's multi-view consistency. To that end, we design a novel three-stage approach. We first use an enhanced Large Reconstruction Model~\cite{hong2023lrm} for single view NeRF prediction. The predicted NeRF captures the coarse geometry and color of the person but lacks details. We then employ a novel-view guided reconstruction stage to improve the overall quality of the person. Specifically, we leverage the generative capability of diffusion models to hallucinate high resolution novel views of the person. Outputs from the first stage contain geometry and appearance information, and are used as conditioning to ensure multi-view consistency during diffusion. Lastly, the novel-view generations are used to guide the higher-fidelity geometry and appearance prediction.
Since our method does not require any human template, we can easily scale up training by including multi-view human datasets to achieve improved generalization capability.
Additionally, unlike existing models that predicts appearance in a deterministic way, \Ours{} leverages the generative power from diffusion models to achieve higher-quality reconstruction.

Our contributions can be summarized as below:
\begin{itemize}
    \item We introduce \Ours{}, a feed-forward model for reconstructing humans with detailed geometry and appearance quality from a single image. Being trained on an extensive dataset (more than 10K identities) including both multi-view RGB data and 3D scans, our model attains substantially enhanced generalizability and excels across a wider spectrum of scenarios.
    \item As the core of our method, we propose a conditional diffusion model for generating high-resolution novel views. Raw renderings from single-view LRM serve as spatial condition to provide geometry and appearance prior. Additional reference networks help preserve the identity of the person.
    The generated novel views contain rich details, and are effective in guiding the final reconstruction of the human with high-fidelity geometry and appearance.
    \item We perform extensive comparisons to existing single-view human reconstruction works ~\cite{saito2019pifu, zheng2021pamir, xiu2022icon, xiu2023econ, yang2023d, zhang2024global, zhang2023sifu, zhang2024global, saito2020pifuhd, hong2023lrm,he2021arch++,liu2021liquid,richardson2023texture,qian2023magic123,corona2022structured,albahar2023humansgd}. We show that \Ours{} outperforms previous methods significantly on a comprehensive evaluation set.
\end{itemize} 

%% file: sec/2_related_work.tex
\section{Related Work}
\textbf{Parametric reconstruction.} Many 3D human reconstruction works ~\cite{li2022cliff,feng2021collaborative,zhang2021pymaf,kanazawa2018end} are built on mesh-based parametric body models, e.g., SMPL~\cite{loper2023smpl}. Given an input image, these methods, referred as Human Mesh Recovery (HMR), employ neural networks to predict the SMPL shape and pose parameters from which the target human body mesh is constructed. This SMPL-conditioned approach greatly reduces the network output complexity and also can be adapted for weakly-supervised training with 2D pose estimates via differentiable mesh rasterization~\cite{kanazawa2018end,weng2022domain}. As SMPL models minimally-clothed human bodies with a smooth mesh of fixed topology, it prevents these methods from reconstructing detailed geometry and texture. Nevertheless, the predicted SMPL mesh is a very good proxy for the fully clothed reconstruction as it captures the base body shape and depicts its pose structure. The promise of HMR motivates follow-up works to predict 3D offsets~\cite{alldieck2019learning,pons2017clothcap,ma2020learning,zhu2019detailed} or build another layer of geometry on top of the base body mesh to accommodate clothed human shapes~\cite{jiang2020bcnet,bhatnagar2019multi}. However, this ``body+offset" strategy lacks the flexibility to represent a wide-range of clothing types. \\
\textbf{Implicit reconstruction.} Implicit-functions offer a topology-agnostic representation for modeling human shapes. PiFU \cite{saito2019pifu} uses pixel-aligned image features to predict 3D occupancy values and colors from sampled 3D points in a predefined grid. Building on this, PIFuHD \cite{saito2020pifuhd} develops a high-resolution module to predict geometric and texture details with additional front-back normal maps as input. While producing expressive reconstruction results for simple inputs like standing humans against clean background, such models are not able to generalize well to in-the-wild scenarios and often yield broken and messy shapes on challenging poses and lightings due to their limited model capacity and lack of a holistic representation. \\
\textbf{Hybrid reconstruction.} An emerging type of approach leverages parametric body models (e.g. SMPL \cite{loper2023smpl}) to improve the generalizability of fully-supervised implicit reconstruction methods.
Starting from a given image and an estimated SMPL mesh, ICON~\cite{xiu2022icon} regresses shapes from locally-queried features to generalize to unseen poses. Wang \etal~\cite{wang2023complete} extends ICON with a GAN-based generative component. ECON~\cite{xiu2023econ} leverages variational normal integration and shape completion to preserve the details of loose clothing. D-IF~\cite{yang2023d} additionally models the uncertainty of the occupancy through an adaptive uncertainty distribution function. GTA~\cite{zhang2024global} uses a hybrid prior fusion strategy combining 3D spatial and SMPL prior-enhanced features. SIFU~\cite{zhang2023sifu} further enhance the 3D features using side-view conditioned features. All these methods leverage SMPL prior and although the incorporation of SMPL does enhance generalizability to large poses, these methods are also constrained by the accuracy of SMPL predictions. Any errors in the estimated SMPL parameters have a cascading effect on the subsequent mesh reconstruction stage. \\
\textbf{Human NeRFs.} 
Neural Radiance Fields (NeRF) \cite{mildenhall2021nerf} marks a pivotal milestone in 3D reconstruction. NeRF empowers the learning of a 3D representation of an object solely from 2D observations. While there exist several notable works that focus on reconstructing human NeRF, these efforts often center around the single video \cite{weng2022humannerf} or image \cite{weng2023zeroavatar,huang2023tech} fine-tuning setting at the cost of substantial computational time, ranging from tens of minutes to hours. In contrast, our focus lies on a feed-forward paradigm that radically reduces the time required for a model to predict a human NeRF from a single image, typically in mere seconds.
A few recent works \cite{kwon2021neural,gao2022mps} also employ a feed-forward paradigm for generalizability, utilizing SMPL as a geometric prior and aggregating features from sparse observations, yet they necessitate multiple views.
A closer related work \cite{hu2023sherf} considers feed-forward human NeRF prediction from a single image. Nonetheless, their method replies on ground truth SMPL body meshes that limit their model representation power.
Our method is completely template-free, 
making NeRF-based human reconstruction more accessible and practical for various scenarios. \\
\textbf{Diffusion-based novel view synthesis.} Many recent works leverage diffusion models for novel view synthesis \cite{liu2023zero,shi2023zero123plus,watson2022novel,liu2023syncdreamer,szymanowicz2023viewset,chan2023generative,wu2023reconfusion}. Maintaining multi-view consistency in geometry and colors for the generated images remains a challenge. To improve multi-view consistency, Zero123++~\cite{shi2023zero123plus} uses reference attention to preserve the global information from the input image. SSDNeRF~\cite{chen2023single}, Viewset Diffusion~\cite{szymanowicz2023viewset} and SyncDreamer~\cite{liu2023syncdreamer} model the joint probability distribution of multi-view images. GNVS~\cite{chan2023generative} and ReconFusion~\cite{wu2023reconfusion} use predicted 3D latent or renderings as conditioning to the diffusion model. 
We use the geometry and appearance renderings from a NeRF prediction as well as global information from the input image and triplane features to ensure multi-view consistency. In contrast to works \cite{liu2023zero,shi2023zero123plus,chan2023generative} that focus on novel view synthesis, we also reconstruct the geometry. In contrast to ReconFusion~\cite{wu2023reconfusion}, our method is feed-forward.


%% file: sec/3_method.tex
\begin{figure}[!t]
    \centering
    \includegraphics[width=\textwidth]{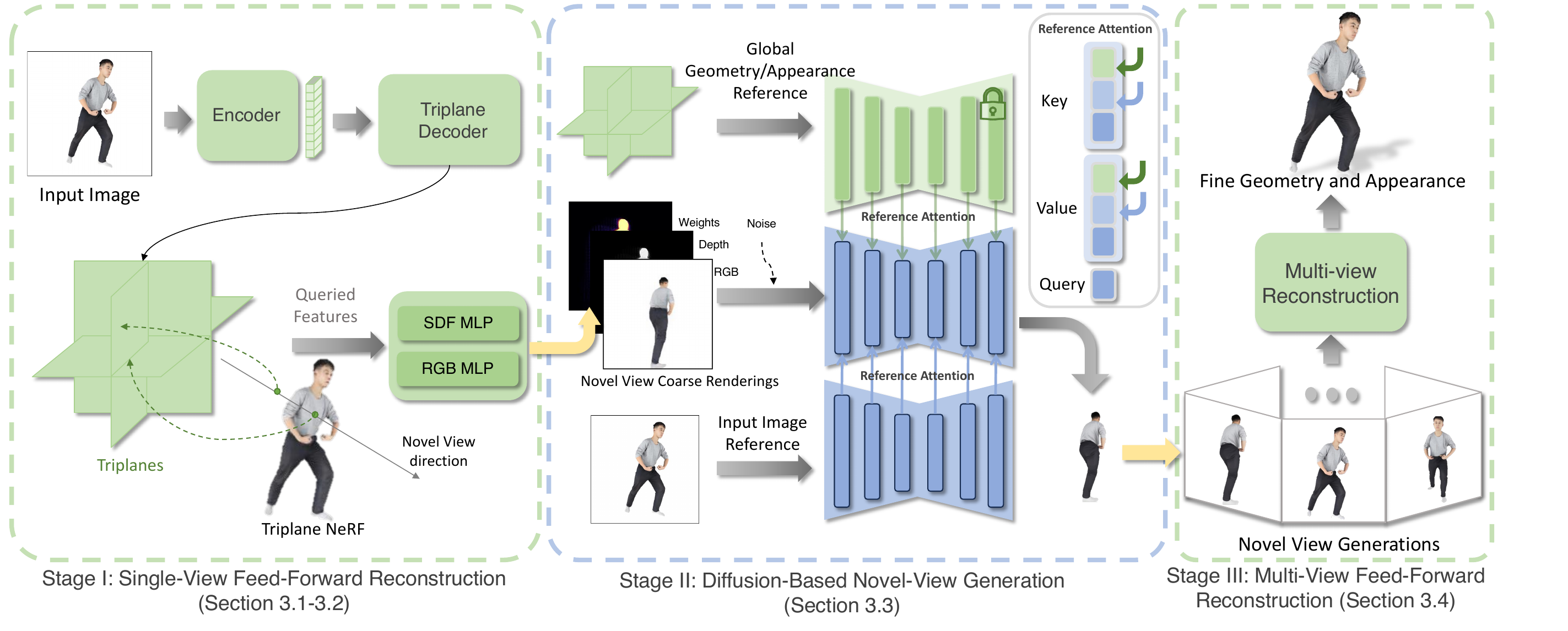}
    \caption{Overview of \Ours{}. Given a single image, we encode the image using ViT \cite{caron2021emerging}, and employ a transformer to decode a triplane representation \cite{chan2022efficient}, followed by SDF and RGB MLPs for volumetric rendering of RGB and depths from novel viewpoints. Next, we use a conditional diffusion model to generate novel-views of the person conditioning on the coarse geometry renderings. From the dense views generated by the diffusion model, we then use a multi-view reconstruction model to generate reconstruction of the person with fine geometry and textures.
    }
    \vspace{-5mm}
    \label{fig:overview}
\end{figure}

\section{Method}
\label{sec:method}
\Ours{} (overview in Figure~\ref{fig:overview}) consists of three stages:
\begin{itemize}
    \item Stage I is built on top of LRM \cite{hong2023lrm} that consists of two building blocks: a transformer-based triplane decoder and triplane NeRF. In Sec.~\ref{sec:method_sv_rec}, we briefly introduce triplane prediction as our model backbone and then in Sec.~\ref{sec:method_render}, we introduce our improved triplane NeRF to enhance the surface reconstruction quality of humans. 
    \item Stage II takes the coarse renderings from the triplane NeRF and uses a specialized diffusion model to generate high-fidelity novel views of the person (Sec.~\ref{sec:method_diffusion}).
    \item Lastly, we reconstruct the person with fine geometry and appearance leveraging the diffused novel views as guide (Sec.~\ref{sec:fine_rec}).
\end{itemize}

\subsection{Single-view Triplane Decoder}
\label{sec:method_sv_rec}
 Given an RGB image $\bx_{\text{input}}$ as input, LRM first applies a pre-trained vision transformer (ViT), DINO \cite{caron2021emerging} to encode the image to patch-wise feature tokens $\{\bh_i \in \mathbb{R}^{768}\}_{i=1}^n $, where $i$ denotes the $i$-th image patch, $n$ is the total number of patches, and 768 is the latent dimension. 

It then uses a transformer module to decode the image tokens into a 3D triplane  \cite{chan2022efficient}. Specifically, the decoder updates learnable tokens to the final triplane features via camera modulation and cross-attention with the image tokens, similar to the design of PerceiverIO \cite{jaegle2021perceiver}. 
More specifically, each transformer layer contains a cross-attention, a self-attention, and a multi-layer perceptron (MLP) sub-layer, where the input tokens to each sub-layer are modulated \cite{peebles2023scalable} by the camera features $\bc$.
The cross-attention layer attends from the triplane features to the image tokens, which can help link image information to the triplane. Then, the self-attention layer further models the intra-modal relationships across the spatially-structured triplane entries. 

Triplane \cite{chan2022efficient} is used as an efficient 3D representation. A triplane $\bT$ contains three axis-aligned feature planes $\bT_{\textrm{XY}}$, $\bT_{\textrm{YZ}}$ and $\bT_{\textrm{XZ}}$. In our implementation, each plane is of dimension $h_T \times w_T \times d_T$ where $h_T\times w_T$ is the spatial resolution, and $d_T$ is the number of feature channels. For any 3D point in the NeRF object bounding box $[-1, 1]^3$ , we can project it onto each of the planes and query the corresponding point features $\bT_{xy}$, $\bT_{yz}$, $\bT_{xz}$ via bilinear interpolation, which is then decoded for rendering (Section \ref{sec:method_render}).

In short, given an input image $\mathcal{I}_1 \in \mathbb{R}^{H\times W\times 3} $, we train an encoder $\E$ and decoder $\D$ s.t. $\{\bh_i\}_{i=1}^n = \E(\mathcal{I}_1)$, and
$\bT_{\textrm{XY}}, \bT_{\textrm{YZ}}, \bT_{\textrm{XZ}} = \D(\{\bh_i\}_{i=1}^n, \bc)$

\subsection{Triplane NeRF}
\label{sec:method_render}
Traditional neural volume rendering methods (as used in LRM \cite{hong2023lrm}) model geometry through a generalized density function. The extraction of this geometry is achieved using a random level set of the density function, which often results in reconstructions that are noisy and of low fidelity. Hence, to improve the fidelity of the reconstructions, we predict Signed Distance Functions (SDF) instead of density. Specifically, we use two MLPs (i.e. ``SDF MLP" and ``RGB MLP" in Figure \ref{fig:overview}) to predict SDF and RGB from the point features queried from the triplane representation $\bT$. 
The SDF MLP takes the point features and output SDF and a latent vector $\bh_{p}$. The RGB MLP takes the point features, latent vector and normals at sampled points $\hat{\bn}_p$ (computed from predicted SDF using finite differences) and output RGB values.
That is, $(\bh_{p}, \text{SDF}) = \text{MLP}_{\text{SDF}} (\bT_{xy}, \bT_{yz}, \bT_{xz})$, $\text{RGB} = \text{MLP}_{\text{RGB}}(\bT_{xy}, \bT_{yz}, \bT_{xz}, \bh_{p}, \hat{\bn}_p )$.
For a ray $r$ emanating from a camera position $\bo$ in direction $\bv \in \mathbb{R}^3$, $||\bv|| = 1$, defined by $\br(t) = \bo + t\bv, t \geq 0$, the color of the corresponding pixel in the rendered image is computed via numerical integration
\vspace{-5mm}
\begin{align}
\small
I(\br) &= \sum_{i=1}^{M} \alpha_i \prod_{i>j} (1-\alpha_j) \text{RGB}_i, \alpha_i = 1 - \textrm{e}^{-\sigma_i \delta_i} \label{eq:nerf}
\end{align}
where $\sigma_i$ is the density converted from SDF using VolSDF~\cite{yariv2021volume}, and $\delta_i$ is the distance between samples. Normals can be rendered using the same formula where we integrate over predicted normals at sampled points instead.\\ 
\textbf{Training objective.} Our training data contains multiple views and their respective camera parameter per human. For each human, we randomly choose a few side views, and render a random $\hat{\bx} \in \mathbb{R}^{h \times w \times 3}$ patch on each view. The ground truth RGB values for the patch is $\bx \in \mathbb{R}^{h \times w \times 3}$. In addition, we render the predicted depths and normals of the patch $\hat{\bd} \in \mathbb{R}^{h \times w \times 3}$ and $\hat{\bn} \in \mathbb{R}^{h \times w}$, and supervise with depths maps $\bd \in \mathbb{R}^{h \times w}$ and normal maps $\bn \in \mathbb{R}^{h \times w \times 3}$. The supervising depth and normal maps can be either ground-truth renderings or off-the-shelf predictions. 
The training objective of our single-view reconstruction method is computed over losses from $V$ rendered views, with the input view as well as $(V-1)$ side views. Overall, the training objective is to minimize $\mathcal{L}$,
\vspace{-5mm}
\begin{align}
\small
\mathcal{L} = &\frac{1}{V} \sum_{v=1}^{V} (\mathcal{L}_{\text{MSE}}(\hat{\bx}_v, \bx_v) \\
& + \lambda_{\text{lpips}} \mathcal{L}_{\text{LPIPS}}(\hat{\bx}_v, \bx_v) \\
& + \lambda_{n} \mathcal{L}_{\text{MSE}}(\hat{\bn}_v, \bn_v)  + \lambda_{d} \mathcal{L}_{\text{DSI}}(\hat{\bd}_v, \bd_v)) \\
& + \lambda_{eik} \mathcal{L}_{\text{Eikonal}}
\end{align}
Subscript $v$ means that the corresponding variable is for the $v_{th}$ supervising view. $\mathcal{L}_\text{MSE}$ is the normalized pixel-wise L2 loss, $\mathcal{L}_\text{LPIPS}$ is the perceptual image patch similarity \cite{zhang2018unreasonable}, and $\mathcal{L}_{\text{DSI}}$ is the scale invariant depth loss \cite{bhat2023zoedepth}. $\mathcal{L}_{\text{Eikonal}}$ is the Eikonal regularization \cite{gropp2020implicit} computed using SDF values of the sampled points along the rays. $\lambda_{\text{lpips}}$, $\lambda_{n}$, $\lambda_{d}$, and $\lambda_{eik}$ are weight coefficients. 

\subsection{Diffusion-Based Novel View Generations}
\label{sec:method_diffusion}
The triplane NeRF from Section~\ref{sec:method_render} captures the geometry and appearance very well through large scale training. However, it often has collapsed reconstruction on the unseen parts, leading to blurry appearance on the back side of the person. To address this limitation, we leverage the power of diffusion models to generate high-fidelity and realistic novel views as guidance for reconstructing the occluded part. Specifically, we condition on the novel-view coarse renderings rendered from the triplane NeRF, and generate high-quality novel views using a diffusion model.

The coarse renderings consist of rendered RGB, depth and weights sum (i.e. sum of the weights in Eqn.~\ref{eq:nerf}). RGB and depth renderings provide the appearance and geometry information from the novel view. We use weights sum as a proxy for the certainty for the rendered content, as parts that are visible often has high weights sum. In other words, weights sum is for the diffusion model to learn to hallucinate the less certain parts.

Using coarse renderings as the only conditioning to the diffusion model often results in generations that do not preserve the identity of the person. Hence, we additionally pass information from the input image and triplanes to the denoiser through reference attention, to make the denoiser be aware of the global information of the person. Reference attention~\cite{zhang2023reference} essentially modifies each of the self-attention layers in the UNet denoiser by concatenating the key and value with that of the reference.

Formally, the objective for diffusion model training is to minimize $\mathcal{L}_{\text{diffusion}}$.
\vspace{-5mm}
\begin{align}
\small
    & r_{\text{coarse}} = (\hat{\bx}_v, \hat{\bd}_v, \hat{\bw}_v)  \\
    \mathcal{L}_{\text{diffusion}} &= \mathbb{E}_{t\sim [1, T]}[|| v - \hat{v}_{\theta}({x}_v^{\text{noised}}; r_{\text{coarse}}, \bx_{\text{input}}, \bT, t) ||^2]
\end{align}
where $\hat{v}_{\theta}$ is a UNet~\cite{ronneberger2015u} (with trainable weights $\theta$) that does $v$-prediction. ${x}_v^{\text{noised}}$ is the noised GT view, $\bx_{\text{input}}$ is the input image, and $\bT$ is the triplanes.

\subsection{Novel-View Guided Feed-Forward Reconstruction}
\label{sec:fine_rec}
Once we obtain dense novel view generations of the human as well as their viewpoints from Sec.~\ref{sec:method_diffusion},
we use a multi-view reconstruction model to reconstruct the human. The multi-view model is trained on the same data with the same objective as the single-view model in Sec.~\ref{sec:method_sv_rec}. In contrast to the single-view model, the multi-view model incorporates camera conditioning within the ViT encoder through modulation~\cite{peebles2023scalable}. We refer readers to LRM~\cite{hong2023lrm} for more details regarding the camera modulation. The triplane decoder in the multi-view model maintains the same architecture as the single-view model, with the exception that it does not take camera conditioning, as it is moved to ViT encoder.

%% file: sec/4_experiments.tex
\section{Experiments}
\textbf{Training data.}
Our complete training set consists of 1,426 high-quality scans (500 from THuman 2.0 \cite{tao2021function4d} and 926 from Alloy++), as well as around 8,000 posed multi-view captures from HuMMan \cite{cai2022humman} v1.0. THuman 2.0 and HuMMan both contain adults with simple clothing. Thus, to further evaluate the generalization capability, we collect Alloy++ from Human Alloy~\cite{humanalloy} and our internal capture. Each scan from Human Alloy has around 40K polygons and our internal capture, 100K polygons. The quality of those scans are similar to that of RenderPeople \cite{renderpeople2018} (100K polygons). Alloy++ contains humans with more challenging clothing, poses, as well as little kids. \\
\textbf{Evaluation sets.} We evaluate on 20 humans from THuman 2.0 and 20 humans from Alloy++, each with renderings from 3 evenly spaced viewpoints. In addition, we create an evaluation set from X-Human \cite{shen2023xavatar}. We randomly sample 2 frames per sequence, which results in 460 frames from 20 human subjects, all with distinct poses. The X-Human testset serves as an out-of-domain evaluation set as none of the models have seen images from this dataset during training. \\
\textbf{Data preprocessing.} For each scan from THuman 2.0 and Alloy++, we center it by the origin and scale them so the longest side has length 1.8. We render each human scan from 32 randomly sampled viewpoints with the same camera pointing toward the origin. For HuMMan v1.0, there are 10 cameras per pose. In total, there are 16K, 14K, and 80K distinct input images from the training split of THuman 2.0, Alloy++ and HuMMan v1.0, respectively. \\
\textbf{Implementation details.}
We train our coarse single-view model and multi-view reconstruction model with 16 A100 GPUs for 7 days. We use batch size of 4 per GPU. During training, we sample rays on a random 64 by 64 patch. We use importance sampling where we sample 48 coarse and 64 fine samples along the rays. $\lambda_{lpips} = 2$, $\lambda_{d}=\lambda_{n}=1$, $\lambda_{eik}=0.5$. We use cosine learning rate scheduler with initial learning rate of $2e-5$. In SDF-density conversion, we use scheduled hyperparameters following \cite{yariv2023bakedsdf} for stable optimization. For diffusion model, we initialize from the Stable Diffusion 2.0 UNet which was trained with v-prediction, and adapt the input layers for our input dimension. We then finetune the UNet on a single A6000 GPU with batch size of 4 and learning rate of $5e-5$. We use a separate UNet for triplane conditioning, which was initialized from the same model. The input layer is updated and rest of the models are frozen. During inference, we use 100 steps of Ancestral sampling with Euler method steps.\\
\textbf{Inference time.}
It takes about 0.7 second for the image encoder and triplane decoder to get the triplane representation from the input image(s), and 1.3 seconds to render a 256 by 256 image from the triplanes. Sampling from diffusion model takes about 5 seconds on a single A6000 GPU. 

\input{tables/main_table_separate}

\input{tables/texture_comparison}
\vspace{-5mm}
\begin{figure*}[h]
    \centering
    \includegraphics[width=\textwidth]{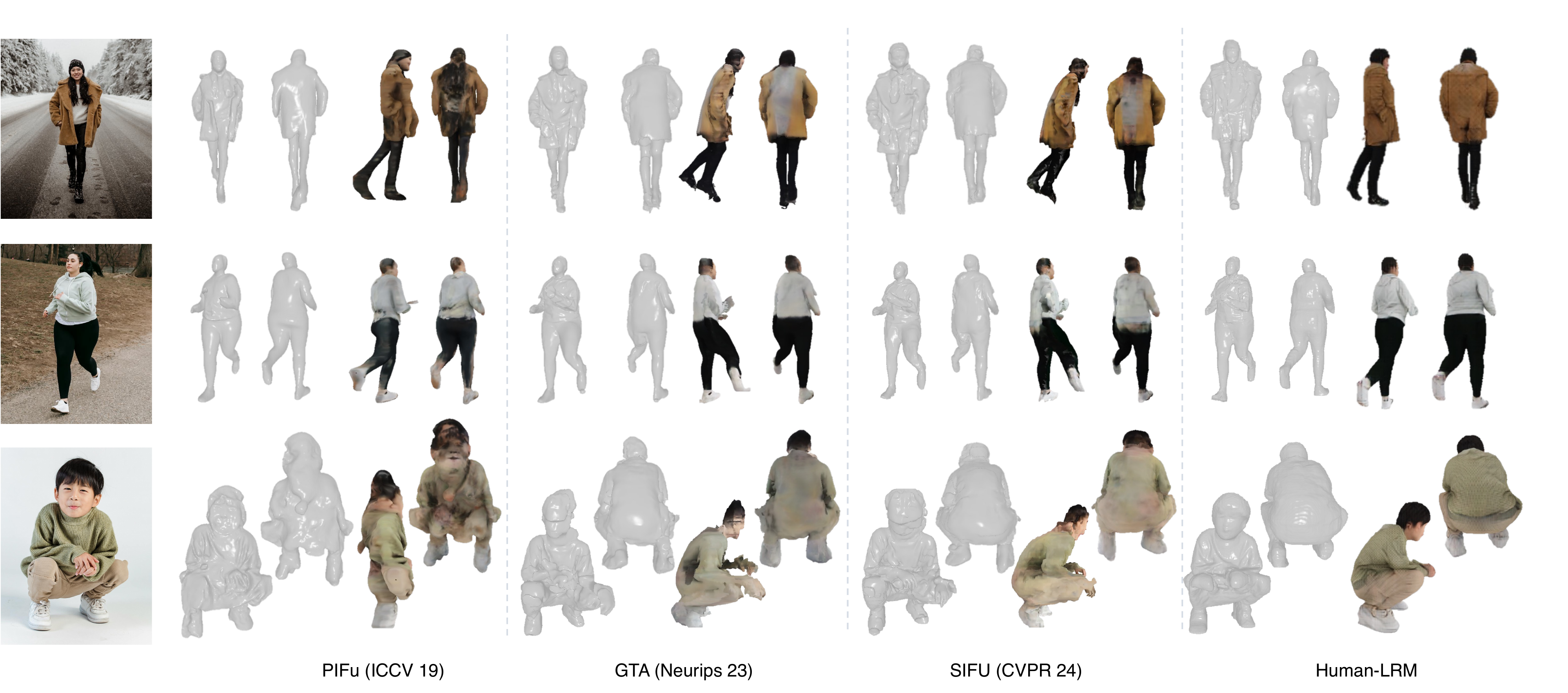}
    \caption{Geometry and appearance comparison with PIFu~\cite{saito2019pifu}, GTA~\cite{zhang2024global} and SIFU~\cite{zhang2023sifu} on in-the-wild images.}
    \label{fig:itw_comparison}
\end{figure*}

\subsection{Geometry Comparisons}
\label{subsec:geom_comparison}
We compare to existing single-view human reconstruction methods PIFu \cite{saito2019pifu}, PIFu-HD \cite{saito2020pifuhd}, Pamir \cite{zheng2021pamir}, ICON \cite{xiu2022icon}, ECON \cite{xiu2023econ}, D-IF~\cite{yang2023d}, GTA~\cite{zhang2024global}, and SIFU~\cite{zhang2023sifu}.
\footnote{
\scriptsize{Wang \etal \cite{wang2023complete} is another related work but we couldn't compare with it as there is no code release and their authors also informed us that their model checkpoints got lost.}}
For works requiring SMPL parameters as input to their model, we use SMPL-X predictions from PIXIE \cite{feng2021collaborative}. To further compare to methods that can work without SMPL, we additionally trained ablated versions of GTA and SIFU by taking out the SMPL-related features.


Following previous works, we report Chamfer distance, Point-to-Surface (P2S) and Normal Consistency (NC). 
First, we compare with their public pretrained models. As some of the baseline methods are trained on the commercially available RenderPeople, we opt for THuman 2.0, a publicly available dataset with a similar scale, to ensure a fair comparison. For GTA and SIFU, we evaluate their released models that are trained on THuman 2.0. We re-train other baselines on the same dataset for fair comparison. 

We report the quantitative results in Table \ref{tab:table1}. ``Human-LRM (Stage I)" is our single-view reconstruction stage as described in Section~\ref{sec:method_sv_rec}. ``Human-LRM - (Full)" is the final reconstruction result following the novel-view guided reconstruction described in Section~\ref{sec:fine_rec}.

\begin{figure*}[h]
    \centering
    \includegraphics[width=\textwidth]{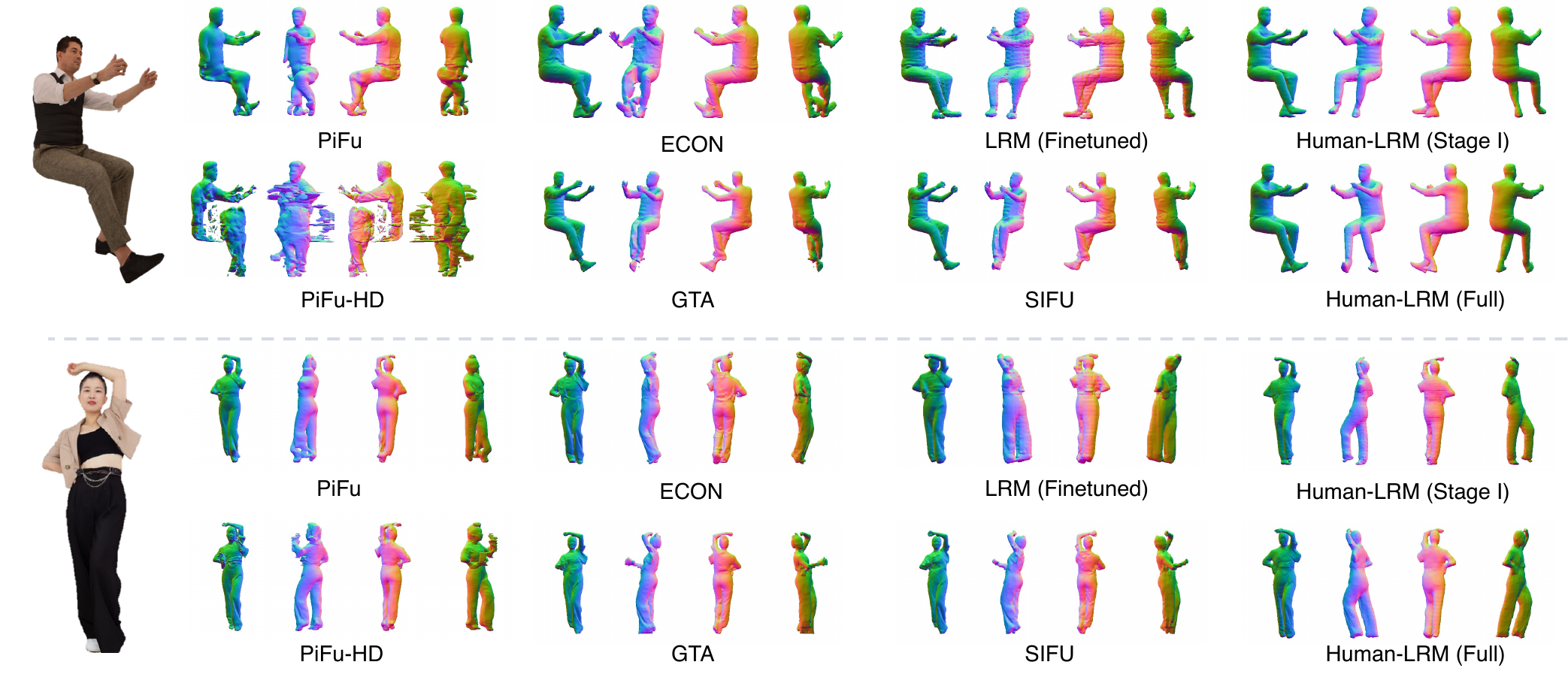}
    \caption{Comparison of our single-view reconstruction model to previous volumetric reconstruction methods: PIFu~\cite{saito2019pifu}, PIFu-HD~\cite{saito2020pifuhd}, ECON~\cite{xiu2023econ}, LRM~\cite{hong2023lrm}, GTA~\cite{zhang2024global}, and SIFU~\cite{zhang2023sifu}. All models are trained on THuman 2.0. For each example we show the geometry (colored by vertex normals) from 4 views.}
    \label{fig:qualitative}
\end{figure*}

As shown by Table \ref{tab:table1}, the geometry predicted by our method consistently outperforms previous works, including works that are prior-free (PIFu, PIFu-HD, SMPL-free GTA) as well as works that require SMPL prior (Pamir, ICON and ECON, GTA, SIFU). The performance of SMPL-guided methods is generally affected by the errors from the predicted SMPL parameters. Even though \cite{xiu2022icon,xiu2023econ,yang2023d,zhang2024global,zhang2023sifu} utilize an optimization algorithm to optimize the SMPL parameters to match the predicted image normals, the errors in the SMPL parameters on some images are still significant. Our method does not rely on a human mesh template such as SMPL and thus does not suffer from this problem. As shown by Fig. \ref{fig:itw_comparison} and Fig. \ref{fig:qualitative}, our method demonstrates exceptional generalizability to challenging cases such as people in difficult poses.
\vspace{-5mm}
\subsection{Appearance Comparisons}
\label{subsec:texture_comparison}
\begin{figure}
    \centering
    \includegraphics[width=0.9\textwidth]{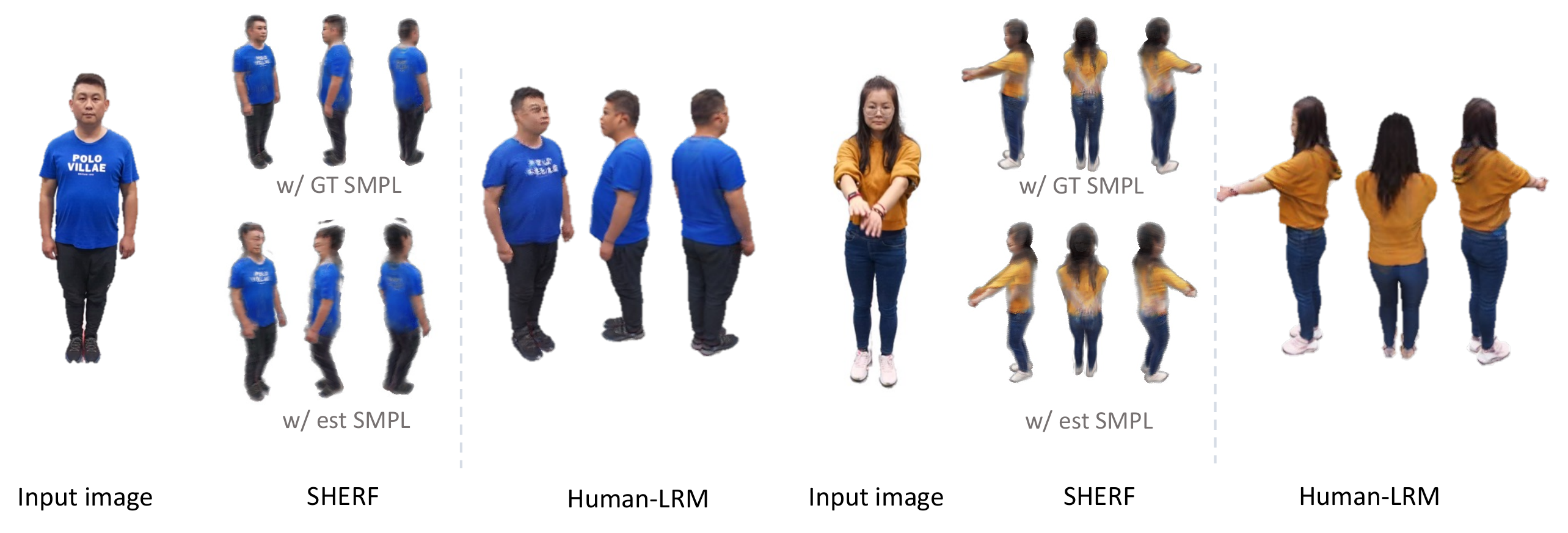}
    \caption{Novel view renderings results on HuMMan v1.0. }
    \label{fig:sherf}
\end{figure}

\paragraph{Comparison to volumetric reconstruction methods.}
A few volumetric reconstruction methods (PIFu~\cite{saito2019pifu}, PIFu-HD~\cite{saito2020pifuhd}, GTA~\cite{zhang2024global}, and SIFU~\cite{zhang2023sifu}) consider joint reconstruction of geometry and color. We compare to those works in Table~\ref{tab:texture} and show that the visual quality from our final mesh renderings are better than previous methods, including the works that use diffusion models. \footnote{
\scriptsize{We are not able to compare to PIFu-HD~\cite{saito2020pifuhd}'s appearance as their color inference module is not released. }} Qualitatively (Fig.~\ref{fig:itw_comparison}), \Ours{} produces much better appearances, especially on the occluded part. Better geometry quality also helps with better color prediction, as shown by the 3rd row.
\vspace{-3mm}
\paragraph{Comparison to generalizable Human NeRF methods.}
Generalizable Human NeRF works (e.g. NHP \cite{kwon2021neural}, MPS-NERF \cite{gao2022mps}, SHERF~\cite{hu2023sherf}) focus is rendering quality instead of geometry. These works assume access to GT SMPL parameters, which is impractical in in-the-wild scenarios. With estimated SMPL parameters their rendering quality declines significantly. For quantitative comparison to the SoTA method SHERF~\cite{hu2023sherf}, we train our model on HuMMan v1.0 \cite{cai2022humman} and evaluate the quality of novel view renderings by SSIM, PSNR and LPIPS following their evaluation protocol. Compared to SHERF with estimated SMPL parameters, SSIM, PSNR and LPIPS from Human-LRM are 30$\%$ higher (better), 10$\%$ higher (better), 45$\%$ lower (better), respectively.  (Full table can be found in \textit{Supplementary}.) The huge performance gap can be attributed to the fact that SHERF's pixel-aligned feature extraction relies heavily on the pixel alignment between the SMPL vertices and RGB. In contrast, our model does not rely on a pose prior, which makes it more resilient and adaptable for in-the-wild scenarios. This robustness is not only demonstrated through notably improved quantitative results but also through the qualitative results illustrated in Fig. \ref{fig:sherf}. In addition, the geometry quality from \Ours{} is significantly better than SHERF, as shown in Fig.~\ref{fig:intro_fig} (c).
\vspace{-5mm}
\vspace{-4mm}
\subsection{Ablations}
\subsubsection{Ablations for single-view reconstruction method (Stage I).}
\begin{figure}
    \centering
    \includegraphics[width=0.6\columnwidth]{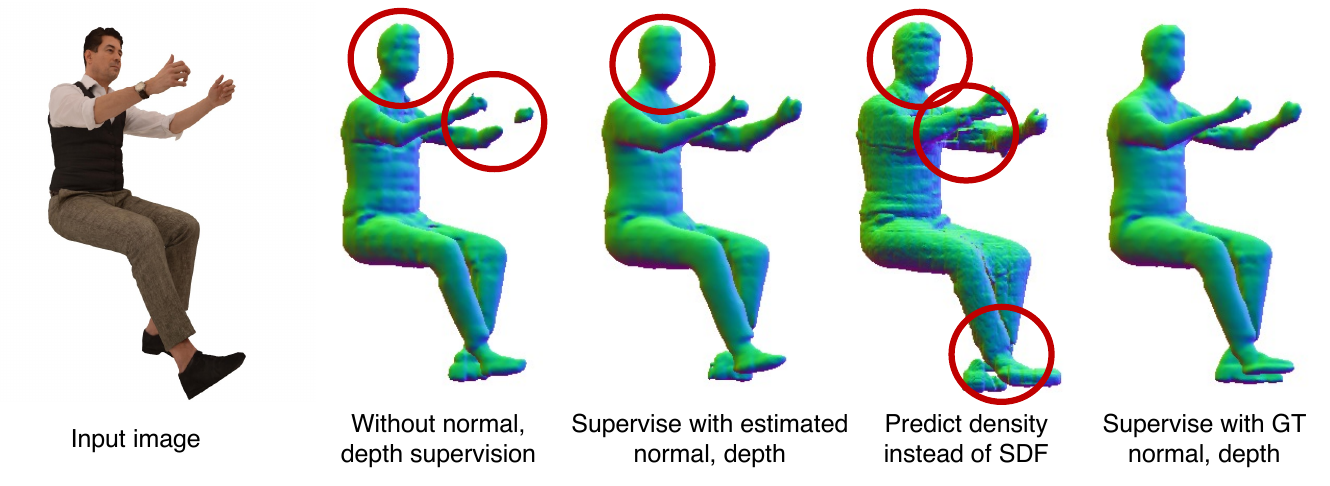}
    \caption{Ablations for the single-view reconstruction model (Stage I). We show the effect of using estimated vs. ground truth normal and depth as supervision as well as using a simple MLP as in LRM \cite{hong2023lrm} to predict the density instead of SDF. }
    \vspace{-5mm}
    \label{fig:ablations}
\end{figure}

\paragraph{Effect of GT normal and depth.} Although for our best model, we do use ground truth normal and depth maps from geometry, we can replace them with estimated ones to achieve comparable performance. We additionally experiment with supervising with predicted normals and depths from off-the-shelf predictors. While there is no notable drop in quantitative performance (Table \ref{tab:ablations}), the surface details are better with GT normal and depth supervision (Figure \ref{fig:ablations}).
\vspace{-3mm}
\paragraph{Effect of predicting SDF.} LRM \cite{hong2023lrm} uses a single MLP to predict the density ($\sigma$) and RGB, followed by NeRF volumetric rendering (i.e. experiment ``predict $\sigma$" in Table~\ref{tab:ablations}). We noticed that the geometry predicted by this approach tended to be more rudimentary in detail (Figure \ref{fig:ablations}). 
\vspace{-3mm}
\paragraph{Effect of the scale of training data.} To further showcase the increased generalization ability of our method with more training data, we train with additional training data from Alloy++ (926 scans). As shown in Table \ref{tab:ablations}, our model shows the best performance, and this performance continues to enhance as we incorporate additional training data. This underscores the significance of expanding the scale of model training to improve generalizability.
\vspace{-5mm}
\vspace{-5mm}
\begin{figure}
    \centering
    \includegraphics[width=\textwidth]{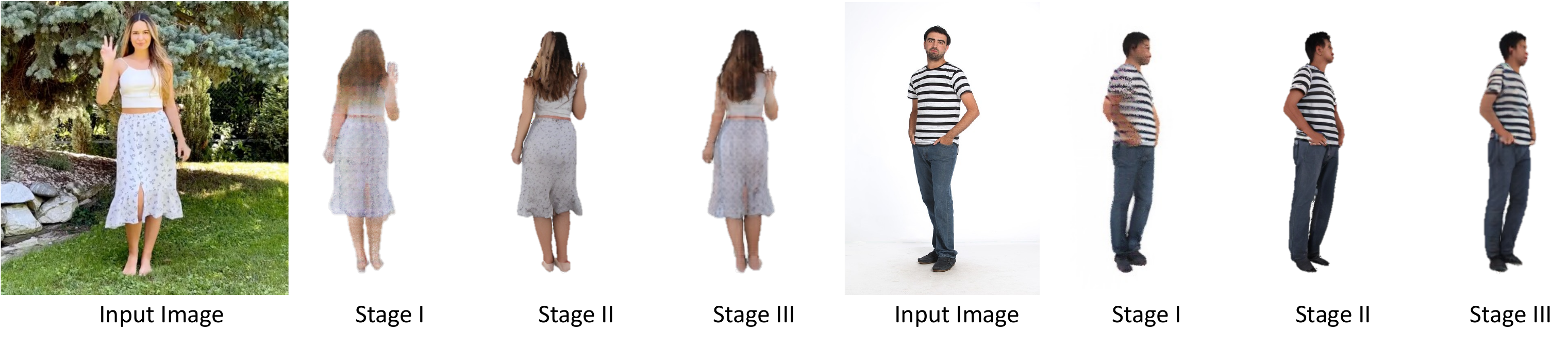}
    \caption{Example novel view results after each stage. Results for Stage I and Stage III are mesh renderings. Results for Stage II are diffusion model outputs (i.e. images). }
    \label{fig:ablations_stages}
\end{figure}
\subsubsection{Ablations for diffusion-guided reconstruction model (Stage II-III).}
\vspace{-3mm}
\paragraph{Effect of number of novel views.} 
We experiment with varying number of novel views during Stage II and evaluate the final geometry quality. For each experiment, we generate novel views from viewpoints evenly spaced around the 0-elevation level of the human. As shown in Table~\ref{tab:ablations_stage2}, with more views during Stage II, our final reconstruction gets increasingly accurate. We report final results (Table~\ref{tab:table1}) using 12 views as reconstruction guide.
\vspace{-3mm}
\paragraph{Effect of diffusion model conditioning.} In Table~\ref{tab:ablations_stage2} we investigate the effect of the each diffusion model conditioning. As shown, each one of the proposed conditioning is useful in improving the generation quality and final geometry quality.
\vspace{-7mm}
\input{tables/ablation_stage1}
\input{tables/ablation_stage2}

%% file: tables/main_table_separate.tex
\begin{table}[h]
\centering
\caption{Geometry comparison to existing single-view reconstruction methods. \textbf{For all models requiring SMPL, PIXIE~\cite{feng2021collaborative} is used for SMPL-X estimation.} }
\vspace{-5mm}
\begin{subtable}[t]{0.75\textwidth}
\centering
\subcaption{Fair comparison of all methods on THuman 2.0 (with the same train-test split). To further compare with SoTA methods that can work without SMPL, we additionally trained ablated GTA model (``w/o SMPL") taking out the SMPL-related features. ($\dagger$ We have verified with the authors of GTA~\cite{zhang2024global} and SIFU~\cite{zhang2023sifu} that the evaluation on THuman 2.0 is correct.) }
\vspace{-2mm}
\resizebox{\textwidth}{!}{%
\begin{tabular}{cc|ccc|ccc|ccc}
\hline
\rowcolor{lightgray} & & \multicolumn{3}{c}{THuman 2.0}& \multicolumn{3}{c}{Alloy++} & \multicolumn{3}{c}{X-Human} \\
\rowcolor{lightgray} Model& SMPL - Free \;&\; Chamfer $\downarrow$ &  P2S $\downarrow$ &  NC $\downarrow$ \;&\;  Chamfer $\downarrow$ &  P2S $\downarrow$ & NC $\downarrow$ \;&\; Chamfer $\downarrow$ & P2S $\downarrow$ &NC $\downarrow$ \\
\hline
PIFu~\cite{saito2019pifu} & \textcolor{black}{\ding{51}}& 1.59 & 1.53 & 0.088 & 1.89 & 1.64 & 0.099 & 1.92 & 1.89 & 0.112 \\
Pamir~\cite{zheng2021pamir}&\textcolor{red}{\ding{55}} & 3.39 & 3.42 & 0.129 & 2.90 & 3.02 & 0.112 & 2.78 & 2.62 & 0.102 \\
ICON~\cite{xiu2022icon}&\textcolor{red}{\ding{55}} & 3.26 & 3.14 & 0.128 & 2.34 & 2.26 & 0.102 & 2.24 & 2.32 & 0.104 \\
ECON~\cite{xiu2023econ}&\textcolor{red}{\ding{55}} & 3.48 & 3.21 & 0.114 & 2.42 & 2.31 & 0.106 & 2.28 & 2.34 & 0.109 \\
D-IF~\cite{yang2023d} & \textcolor{red}{\ding{55}} & 3.34 & 3.75 & 0.126 & 2.49& 2.52& 0.118& 2.32& 2.47&0.109\\
GTA~\cite{zhang2024global} $\dagger$  & \textcolor{red}{\ding{55}} & 3.18 & 3.86 & 0.131 & 2.26& 2.92& 0.134 & 2.22& 2.13&0.108\\
SIFU~\cite{zhang2023sifu} $\dagger$  & \textcolor{red}{\ding{55}} & 2.66 & 3.63 & 0.112 & 2.68& 2.46& 0.109& 2.10& 2.08&0.083\\
GTA~\cite{zhang2024global} retrained w/o SMPL& \textcolor{black}{\ding{51}}& 1.84 & 1.66 & 0.089& 1.92& 1.73& 0.091& 1.89& 1.90&0.080\\
Human-LRM (Stage I)& \textcolor{black}{\ding{51}}& \cellcolor[HTML]{ffbf69} 1.27& \cellcolor[HTML]{ffbf69}1.41& \cellcolor[HTML]{ffbf69} 0.074& \cellcolor[HTML]{ffbf69} 1.58 & \cellcolor[HTML]{ffbf69} 1.48 & \cellcolor[HTML]{ffbf69} 0.072 & \cellcolor[HTML]{ffbf69} 1.23 & \cellcolor[HTML]{ffbf69} 1.21 & \cellcolor[HTML]{ffbf69} 0.065 \\
\textbf{Human-LRM (Full)} & \textcolor{black}{\ding{51}}& \cellcolor[HTML]{ff9f1c} 1.26 & \cellcolor[HTML]{ff9f1c} 1.37 & \cellcolor[HTML]{ff9f1c} 0.071 & \cellcolor[HTML]{ff9f1c} 1.55 & \cellcolor[HTML]{ff9f1c} 1.43 &\cellcolor[HTML]{ff9f1c} 0.067 & \cellcolor[HTML]{ff9f1c} 1.21 & \cellcolor[HTML]{ff9f1c} 1.20 & \cellcolor[HTML]{ff9f1c} 0.064 \\
\hline
\end{tabular}
}
\end{subtable}

\vspace{2mm}

\begin{subtable}[t]{\textwidth}
\centering
\subcaption[]{Comparison of off-the-shelf models. The size of each training set is \textit{italicized}. }
\vspace{-2mm}
\resizebox{\textwidth}{!}{%
\begin{tabular}{ccc|ccc|ccc|ccc}
\hline
\rowcolor{lightgray} \;&\;   & & \multicolumn{3}{c}{THuman 2.0}& \multicolumn{3}{c}{Alloy++} & \multicolumn{3}{c}{X-Human} \\
\rowcolor{lightgray} Model \;&\; Training Data  & SMPL - Free \;&\; Chamfer $\downarrow$ &  P2S $\downarrow$ &  NC $\downarrow$ \;&\;  Chamfer $\downarrow$ &  P2S $\downarrow$ & NC $\downarrow$ \;&\; Chamfer $\downarrow$ & P2S $\downarrow$ &NC $\downarrow$ \\
\hline
PIFu \cite{saito2019pifu} & RenderPeople-\textit{442}  & \textcolor{black}{\ding{51}}& 2.13& 1.97 & 0.134&  2.24& 2.19& 0.120& 1.96& 1.90&0.120\\
PIFu-HD \cite{saito2020pifuhd}& RenderPeople-\textit{450}  &  \textcolor{black}{\ding{51}}& 1.65 & 1.68 & 0.091& 1.94& 1.99& 0.099& 1.92& 2.04&0.123\\
Pamir \cite{zheng2021pamir}& Twindom~\cite{Twindom}-\textit{900} + DeepHuman~\cite{zheng2019deephuman}-\textit{600}  & \textcolor{red}{\ding{55}} & 3.46& 3.23& 0.123& 2.42& 2.38& 0.109& 2.51&2.32&0.107\\
ICON \cite{xiu2022icon}& RenderPeople-\textit{450}  &\textcolor{red}{\ding{55}} & 3.28 & 3.17 & 0.120 & 2.23 & 2.29 & 0.099 & 2.37 & 2.19 & 0.102 \\
LRM \cite{hong2023lrm} & Objaverse~\cite{deitke2023objaverse} + MVImgNet~\cite{yu2023mvimgnet} -\textit{730K in total} & \textcolor{black}{\ding{51}}& 2.36 & 2.08 & 0.109 & 1.74 & 1.82 & 0.083 & 2.21 & 2.09 & 0.103 \\
Human-LRM (Stage I)&  THuman 2-\textit{500}, Alloy++ -\textit{926}, HuMMan v1-\textit{8000} & \textcolor{black}{\ding{51}}& \cellcolor[HTML]{ffbf69}1.26& \cellcolor[HTML]{ffbf69}1.40& \cellcolor[HTML]{ffbf69}0.070& \cellcolor[HTML]{ffbf69}1.58& \cellcolor[HTML]{ffbf69}1.47& \cellcolor[HTML]{ffbf69}0.068& \cellcolor[HTML]{ffbf69}1.34& \cellcolor[HTML]{ffbf69}1.31&\cellcolor[HTML]{ffbf69}0.082\\
\textbf{Human-LRM (Full)} & THuman 2-\textit{500}, Alloy++ -\textit{926}, HuMMan v1-\textit{8000} & \textcolor{black}{\ding{51}}& \cellcolor[HTML]{ff9f1c} 1.24 & \cellcolor[HTML]{ff9f1c} 1.36 & \cellcolor[HTML]{ff9f1c} 0.068 & \cellcolor[HTML]{ff9f1c} 1.52 & \cellcolor[HTML]{ff9f1c} 1.43 & \cellcolor[HTML]{ff9f1c} 0.067 & \cellcolor[HTML]{ff9f1c} 1.20 & \cellcolor[HTML]{ff9f1c} 1.19 & \cellcolor[HTML]{ff9f1c} 0.062 \\
\hline

\end{tabular}
}
\end{subtable}

\label{tab:table1}
\end{table}

%% file: tables/texture_comparison.tex
\begin{table}[h]
    \centering
    \caption{\small{Appearance comparison on THuman 2.0. Baseline numbers are from \cite{zhang2023sifu}.}}
    \resizebox{0.6\textwidth}{!}{%
    \begin{tabular}{cc|ccc}
        \hline
        \rowcolor{lightgray} Model \;&\; Diffusion-based\; &\; PSNR ${\uparrow}$ \;&\; SSIM ${\uparrow}$ \;&\; LPIPS ${\downarrow}$  \\
         \hline
         ARCH++ ~\cite{he2021arch++}& \textcolor{red}{\ding{55}} & 10.6600 & - & - \\
        PIFu~\cite{saito2019pifu} & \textcolor{red}{\ding{55}}
 & 18.0934 & 0.9117 & 0.1372  \\
        Impersonator++~\cite{liu2021liquid} & \textcolor{red}{\ding{55}}
 &  16.4791&  0.9012&   0.1468\\
        TEXTure~\cite{richardson2023texture} & {\ding{51}}
 &  16.7869&  0.8740&   0.1435\\
        Magic123~\cite{qian2023magic123} & {\ding{51}}
 &  14.5013&  0.8768&   0.1880\\
        S3F~\cite{corona2022structured} & \textcolor{red}{\ding{55}}
 &  14.1212&  0.8840&   0.1868\\
        HumanSGD~\cite{albahar2023humansgd} & {\ding{51}}
 & 17.3651& 0.8946& 0.1300\\
        GTA~\cite{zhang2024global} &  \textcolor{red}{\ding{55}} & 18.0500 & - & - \\
        SIFU~\cite{zhang2023sifu} &  \textcolor{red}{\ding{55}}
 & 22.1024 & \cellcolor[HTML]{ffbf69} 0.9236 & 0.0794  \\
 \hline
 Human-LRM (Stage I)& \textcolor{red}{\ding{55}}& \cellcolor[HTML]{ffbf69}22.8902& 0.9207& \cellcolor[HTML]{ffbf69}0.0782\\
        Human-LRM (Full)& {\ding{51}}
 & \cellcolor[HTML]{ff9f1c}24.8050 & \cellcolor[HTML]{ff9f1c} 0.9364 & \cellcolor[HTML]{ff9f1c}0.0604 \\
        \hline
    \end{tabular}
    }
    \label{tab:texture}
\end{table}

%% file: tables/ablation_stage1.tex
\begin{table}[h]
\centering
\caption{Ablations of our single-view reconstruction model (Stage I). Top: Effect of using depth and normal maps (``d.n.") for supervision and predicting SDFs. Bottom: Effect of the scale of training data. }
\resizebox{0.7\textwidth}{!}{
\begin{tabular}{cc|ccc}
\hline
\rowcolor{lightgray} &   & \multicolumn{3}{c}{THuman 2.0 } \\
\rowcolor{lightgray} Model& Training Data  & Chamfer $\downarrow$ &P2S $\downarrow$ \;&NC $\downarrow$\\ \hline
Est. d.n. & THuman 2.0, HuMMan v1.0 & 1.35 & 1.20 & 0.076\\
No d.n. & THuman 2.0, HuMMan v1.0 & 1.34 & 1.20 & 0.075\\
Predict $\sigma$ & THuman 2.0, HuMMan v1.0 & \cellcolor[HTML]{ffbf69}1.28 & \cellcolor[HTML]{ffbf69}1.22 & \cellcolor[HTML]{ffbf69}0.072\\
Full Model & THuman 2.0, HuMMan v1.0 & \cellcolor[HTML]{ff9f1c} 1.25 & \cellcolor[HTML]{ff9f1c}1.20 & \cellcolor[HTML]{ff9f1c}0.069 \\
\hline
Small training set & THuman 2.0  & 1.27 & 1.41 & 0.074 \\
Medium training set & THuman 2.0, HuMMan v1.0 & \cellcolor[HTML]{ffbf69}1.25 & \cellcolor[HTML]{ffbf69}1.20 & \cellcolor[HTML]{ffbf69}0.069 \\
Large training set & THuman 2.0, Alloy++, HuMMan v1.0 & \cellcolor[HTML]{ff9f1c} 1.22 & \cellcolor[HTML]{ff9f1c} 1.02 & \cellcolor[HTML]{ff9f1c} 0.065 \\ \hline
\end{tabular}
}
\label{tab:ablations}
\end{table}

%% file: tables/ablation_stage2.tex
\begin{table}
    \vspace{-10mm}
    \caption{Ablations of diffusion-guided reconstruction model on THuman 2.0.}
    \vspace{-2mm}
    \centering
    \begin{subtable}[t]{0.7\textwidth}
    \centering
    \subcaption{Effect of number of diffused views on the final geometry quality.}
    \vspace{-2mm}
    \resizebox{\textwidth}{!}{%
    \begin{tabular}{c|cc|ccccc}
    \hline
    \rowcolor{lightgray} & & & \multicolumn{5}{c}{Normal Consistency ${\downarrow}$ }\\
         \rowcolor{lightgray} $\#$ of diffused views & Chamfer ${\downarrow}$ \;&\; P2S ${\downarrow}$ \;&\; Overall \;&\; Front \;&\; Right \;&\; Back \;&\; Left\\
         \hline
         2& 1.994& 2.070&
      0.0967&0.0530& 0.1560& 0.0651&0.1119\\
     4& 1.389& 1.485&  0.0753&0.0507& 0.1020& 0.0639&0.0840\\
    \rowcolor[HTML]{ffbf69} 8& 1.284& 1.391&  0.0714&0.0480& \cellcolor[HTML]{ff9f1c} 0.0974& 0.0591&0.0810\\
    \rowcolor[HTML]{ff9f1c} 12 & 1.264 & 1.376 & 0.0707 & 0.0467 & 0.0974 & 0.0576 & 0.0809 \\
    \hline
    \end{tabular}
    }
    \end{subtable}

    
    \begin{subtable}[t]{0.6\textwidth}
    \centering
    \subcaption{Effect of diffusion model conditioning on diffusion model output quality and final geometry quality. }
    \vspace{-2mm}
    \resizebox{\textwidth}{!}{%
    \begin{tabular}{c|ccc|ccc}
    
    \rowcolor{lightgray} & \multicolumn{3}{c}{Diffusion Output Quality} & \multicolumn{3}{c}{Final Geometry Quality}\\ 
         \rowcolor{lightgray} Conditioning & PSNR${\uparrow}$ \;& SSIM${\uparrow} \;$&LPIPS${\downarrow}$\;&Chamfer ${\downarrow}$ \;& P2S ${\downarrow}$ \;& NC ${\downarrow}$ \;\\
         \hline
         w/o depth&    \cellcolor[HTML]{ffbf69}24.926&\cellcolor[HTML]{ffbf69}0.937&\cellcolor[HTML]{ffbf69}0.059&1.399 & 1.674 & 0.0744 \\ 
         w/o weights sum &    24.918&0.936&\cellcolor[HTML]{ffbf69}0.059&1.355 & 1.479 & 0.0753 \\ 
     w/o input image&    23.338&0.931&0.067&1.345& 1.473&  0.0752\\ 
     w/o triplanes&    24.909&\cellcolor[HTML]{ffbf69}0.937&\cellcolor[HTML]{ffbf69}0.059&\cellcolor[HTML]{ffbf69}1.271 & \cellcolor[HTML]{ffbf69}1.390 & \cellcolor[HTML]{ffbf69}0.0712 \\
     Full &    \cellcolor[HTML]{ff9f1c}25.067&\cellcolor[HTML]{ff9f1c}0.938&\cellcolor[HTML]{ff9f1c}0.058&\cellcolor[HTML]{ff9f1c}1.264 & \cellcolor[HTML]{ff9f1c}1.376 & \cellcolor[HTML]{ff9f1c}0.0707 \\
    \hline
    \end{tabular}
    }
    
    \end{subtable}

    \label{tab:ablations_stage2}
\end{table}

%% file: sec/5_conclusion.tex
\section{Conclusion and Future Work}
We introduced a novel approach for reconstructing human NeRFs from a single image. What sets our approach apart from previous implicit volumetric human reconstruction methods is its remarkable scalability, making it adaptable for training on large and diverse multi-view RGB datasets. Additionally, we proposed a coarse-to-fine reconstruction strategy guided by dense novel generations from a diffusion model. The dense novel views serve a strong geometry and texture guide that effectively enhances the overall quality of the final reconstruction.

Although \Ours{} excels in capturing global geometry, it still falls short in preserving finer facial and hand details. Future directions include utilizing more powerful representation than triplanes or additional refinement techniques. 

%% file: sec/6_acknowledgement.tex
\paragraph{Acknowledgement}
The authors would like to thank Chun-Hao Huang for the insightful discussions and valuable suggestions, which significantly contributed to the progression and shaping of this paper.

%% file: sec/X_suppl.tex



\section*{\textit{Supplementary Materials}}

\subsection*{A. Comparison to Generalizable Human NeRF Methods}
\begin{figure}[h]
    \centering
    \includegraphics[width=0.9\columnwidth]{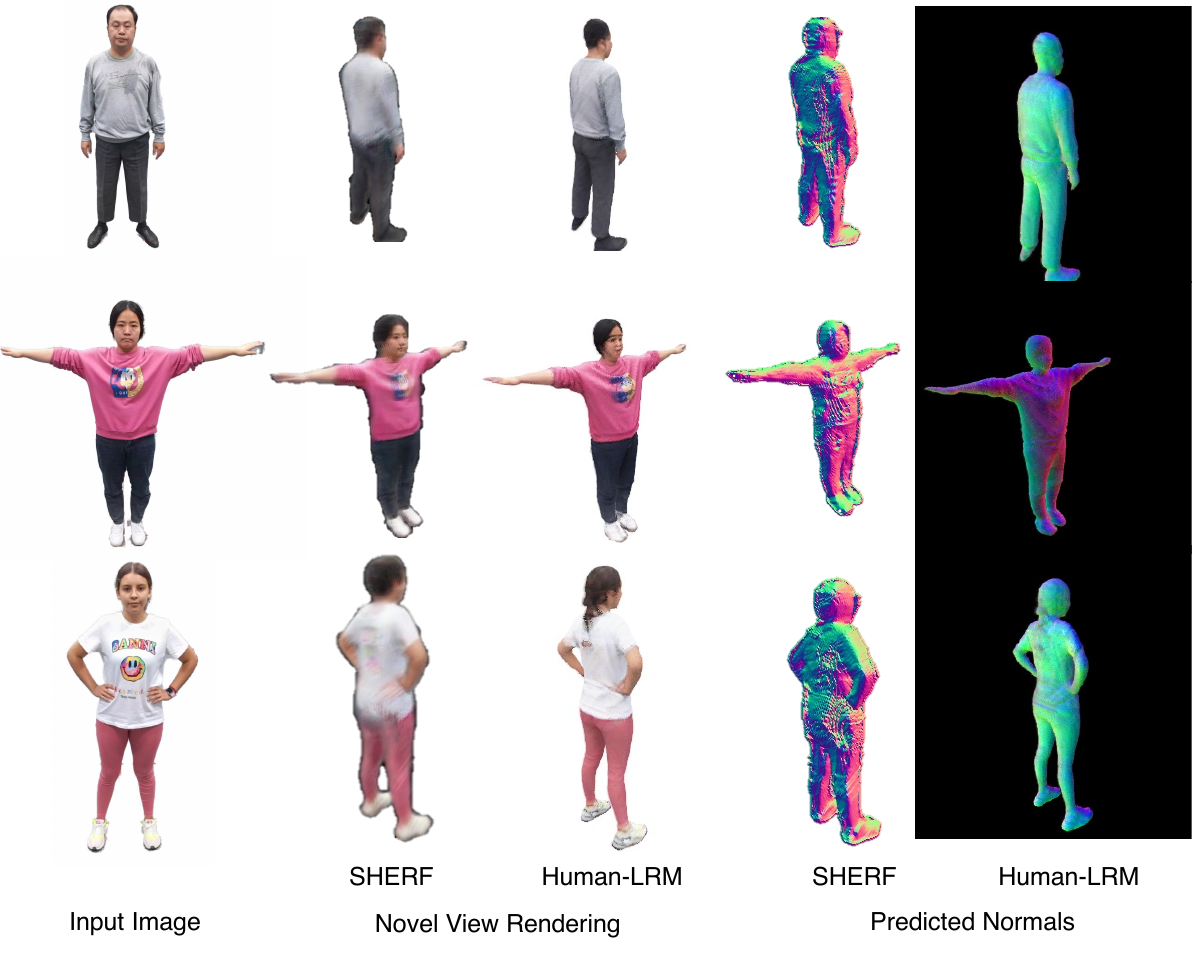}
    \caption{Additional comparison to SHERF.}
    \label{fig:sherf_geom}
\end{figure}

We include additional novel view renderings results from SHERF \cite{hu2023sherf} and \Ours{} on HuMMan v1.0~\cite{cai2022humman} in Figure~\ref{fig:sherf_geom}. We include qualitative comparisons between SHERF's normals (computed from estimated depths) and our normal predictions. As shown, the geometry quality from \Ours{} is significantly better than SHERF.

\input{tables/gen_nerf}

\input{tables/suppl_table}

\begin{figure*}[h]
    \centering
    \includegraphics[width=0.9\textwidth]{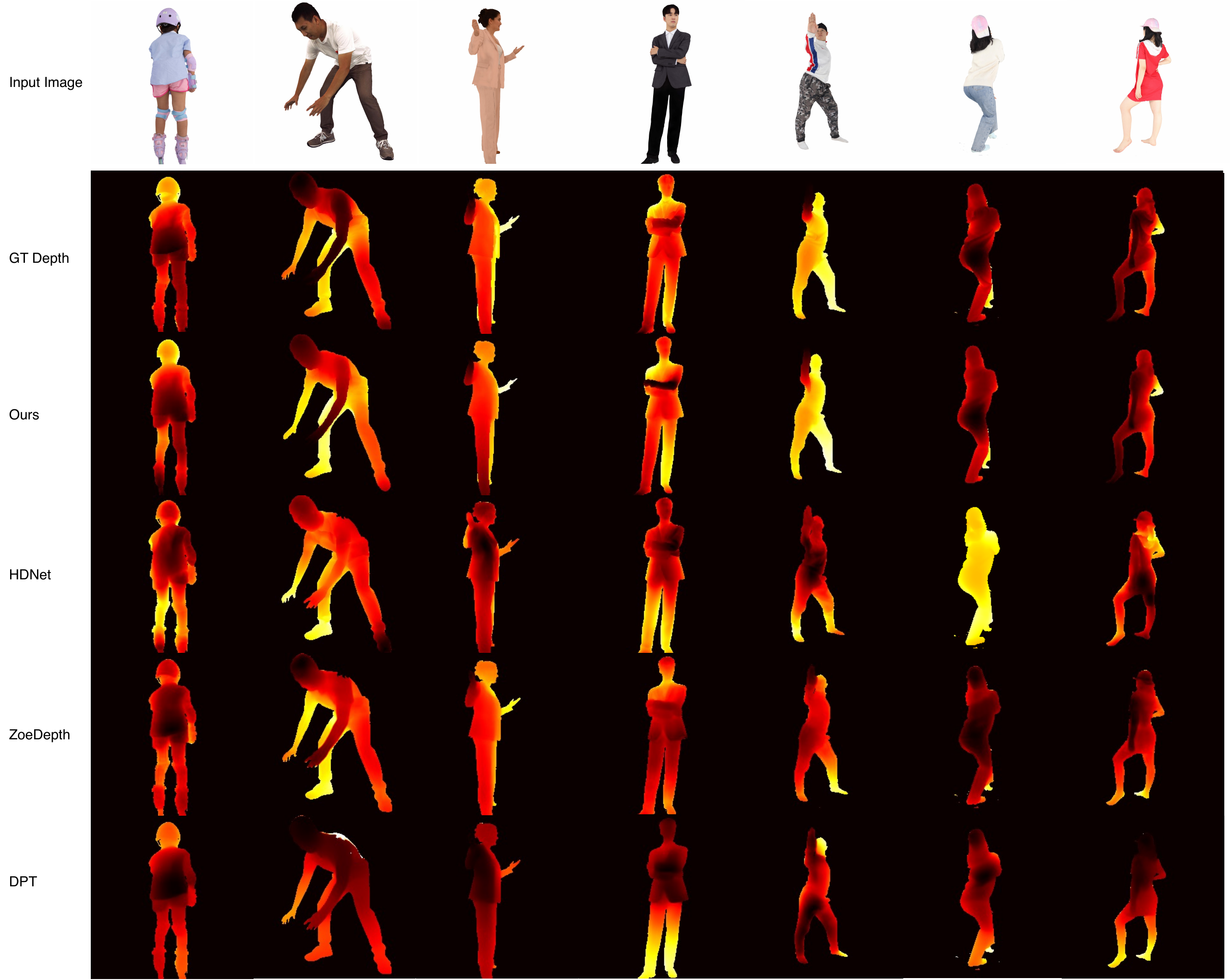}
    \caption{Depth comparison to HDNet~\cite{jafarian2021learning}, ZoeDepth~\cite{bhat2023zoedepth} and DPT~\cite{dpt}. Red color means the region is closer. }
    \label{fig:depth}
\end{figure*}

\begin{figure*}[h]
    \centering
    \includegraphics[width=0.9\textwidth]{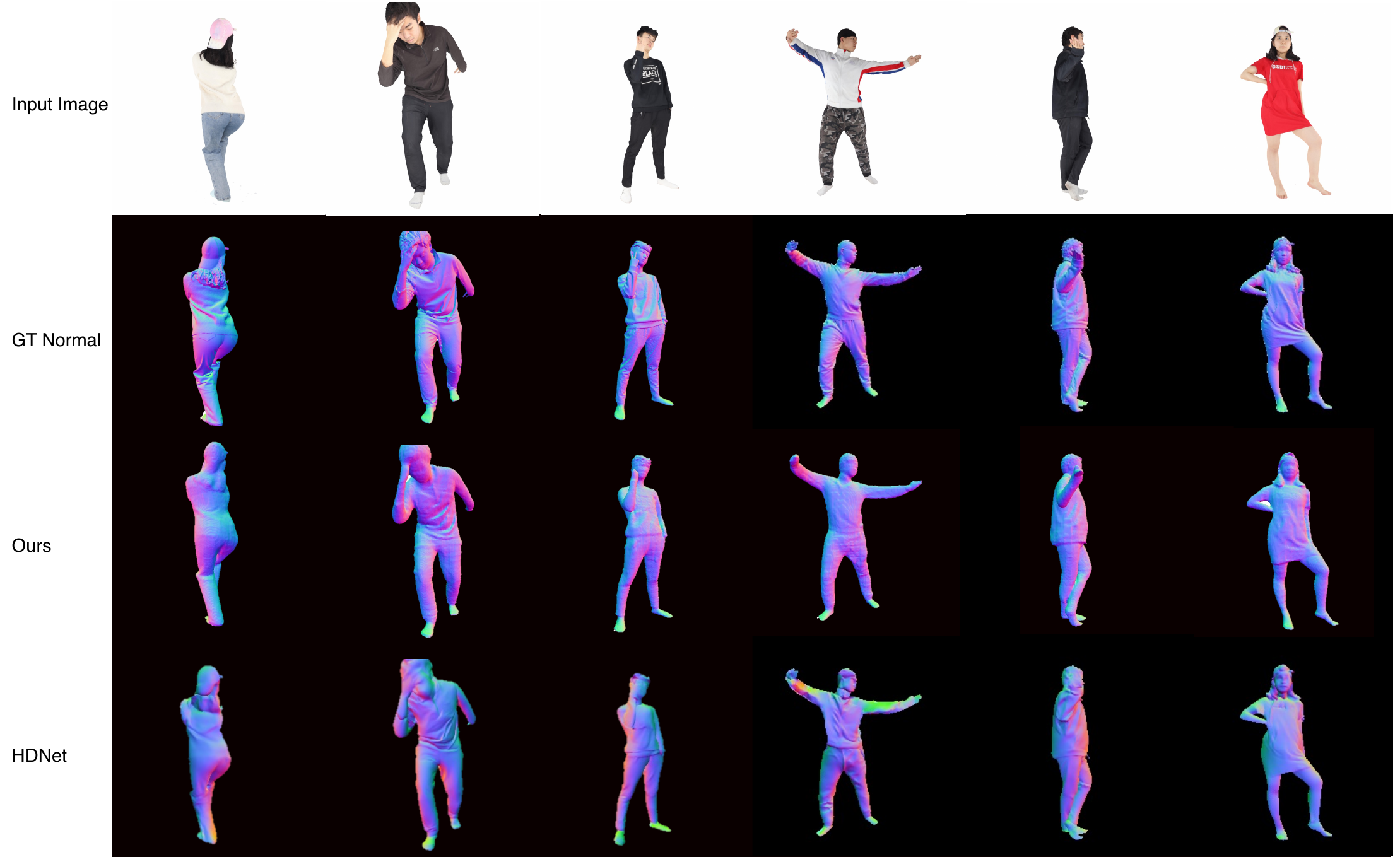}
    \caption{Normal comparison to HDNet~\cite{jafarian2021learning}.}
    \label{fig:normal}
\end{figure*}

\subsection*{B. Comparison to SoTA Depth and Normal Estimation Methods}
We compare the depth quality of our reconstructed geometry to state-of-the-art depth estimation works ZoeDepth \cite{bhat2023zoedepth} and DPT \cite{dpt}. Furthermore, we include a comparison of the depth and normal from our predicted geometry with HDNet \cite{jafarian2021learning}, a method tailored for human surface reconstruction. The results (Table~\ref{tab:hdnet}) demonstrate that our approach surpasses all baseline methods across various datasets. We include qualitative results for depth estimation in Figure~\ref{fig:depth}, and normal estimation in Figure~\ref{fig:normal}. 

\subsection*{C. Personal and human subjects data.} In our experiments, we use public datasets THuman 2.0~\footnote{\url{https://github.com/ytrock/THuman2.0-Dataset}}~\cite{tao2021function4d}, X-Human~\footnote{\url{https://skype-line.github.io/projects/X-Avatar/}}~\cite{shen2023xavatar}, Humman v1.0~\footnote{\url{https://caizhongang.com/projects/HuMMan/}}~\cite{cai2022humman}, and commercially available dataset Alloy~\footnote{\url{https://humanalloy.com}}~\cite{humanalloy}. These datasets are widely used for human reconstruction research and we direct to their respective website for information about their data collection procedures.

\subsection*{D. Ethics statement.} \Ours{} is capable of transforming a single image into a 3D human. While \Ours{}'s results are not yet advanced enough to deceive human perception, it's important to remain vigilant about possible ethical concerns. While we discourage such practices, there is a potential risk that the 3D human models created could be used to produce deceptive content.

%% file: tables/gen_nerf.tex
\begin{table}
\centering
\caption{Comparison to generalizable human NeRF methods on HuMMan v1.0 \cite{cai2022humman}. Top section: Feed-forward methods that use GT SMPL parameters during inference. 
    Bottom section: methods that do not use GT SMPL during inference. }

\resizebox{0.6\textwidth}{!}{%
    \begin{tabular}{cc|ccc}
    \hline
       \rowcolor{lightgray}  Method & GT SMPL & PSNR $\uparrow$&  SSIM $\uparrow$& LPIPS $\downarrow$\\
         \hline
         NHP \cite{kwon2021neural}& \checkmark & 18.99 & 0.84 & 0.18 \\
         MPS-NERF \cite{gao2022mps}& \checkmark & 17.44 & 0.82 & 0.19  \\
         SHERF \cite{hu2023sherf} & \checkmark & \cellcolor[HTML]{ff9f1c} 20.83 &  \cellcolor[HTML]{ff9f1c} 0.89 & \cellcolor[HTML]{ff9f1c} 0.12 \\
         \hline
         SHERF \cite{hu2023sherf} & $\times$ & 14.46 & 0.79 & 0.20 \\
         Ours & $\times$ & \cellcolor[HTML]{ff9f1c} 18.98 & \cellcolor[HTML]{ff9f1c} 0.82 & \cellcolor[HTML]{ff9f1c} 0.11 \\
         \hline
    \end{tabular}
    }
    
    \label{tab:sherf}
\end{table}

%% file: tables/suppl_table.tex

\begin{table*}
\centering
\resizebox{\textwidth}{!}{%
    \begin{tabular}{c|cc|cc|cc}
 & \multicolumn{2}{c}{THuman 2.0}& \multicolumn{2}{c}{Alloy ++}& \multicolumn{2}{c}{X-Human}\\
         Method & Depth Error ($\downarrow$) & Normal Error ($\downarrow$) & Depth Error ($\downarrow$) & Normal Error ($\downarrow$) & Depth Error ($\downarrow$) & Normal Error ($\downarrow$) \\
         \hline
 DPT& 2.44 $\pm$ 0.84& -& 2.91 $\pm$ 1.24& -& 3.17 $\pm$ 1.20&-\\
 ZoeDepth&  2.20 $\pm$ 0.96& -&  2.41 $\pm$ 1.01& -& 2.08 $\pm$ 0.88&-\\
         
         HDNet & 2.27 $\pm$  0.80&  0.58 $\pm$  0.07& 2.37 $\pm$ 1.04& 0.47 $\pm$  0.06& 2.30 $\pm$ 0.83&0.48 $\pm$  0.06\\
         Human-LRM (Ours)& \textbf{0.91} $\pm$  \textbf{0.39} &\textbf{0.38} $\pm$  \textbf{0.04}&   \textbf{1.79} $\pm$  \textbf{0.78} &\textbf{0.41} $\pm$  \textbf{0.05}& \textbf{1.28} $\pm$  \textbf{0.49}& \textbf{0.33} $\pm$  \textbf{0.04}\\
    \end{tabular}
    }
    \caption{Comparison with HDNet \cite{jafarian2021learning}, ZoeDepth \cite{bhat2023zoedepth}, and DPT \cite{dpt}. We report mean and standard deviation ($\pm$) across all test samples.}
    \label{tab:hdnet}
\end{table*}